# Tensor Sparse PCA and Face Recognition: A Novel Approach

Loc Tran
Laboratoire CHArt EA4004
EPHE-PSL University, France
tran0398@umn.edu
Linh Tran, Bao Bui, Trang Hoang
Ho Chi Minh University of Technology, Vietnam
linhtran.ut@gmail.com

Abstract: Face recognition is the important field in machine learning and pattern recognition research area. It has a lot of applications in military, finance, public security, to name a few. In this paper, the combination of the tensor sparse PCA with the nearest-neighbor method (and with the kernel ridge regression method) will be proposed and applied to the face dataset. Experimental results show that the combination of the tensor sparse PCA with any classification system does not always reach the best accuracy performance measures. However, the accuracy of the combination of the sparse PCA method and one specific classification system is always better than the accuracy of the combination of the PCA method and one specific classification system and is always better than the accuracy of the classification system itself.

I.     Introduction

Face recognition is the important field in machine learning and pattern recognition research area. It has a lot of applications in military, finance, public security, to name a few. Computer scientists have worked in face recognition research area for almost three decades. In the early 1990, the Eigenface [1] technique has been employed to recognize faces and it can be considered the first approach used to recognize faces. The Eigenface [1] used the Principle Component Analysis (PCA) which is one of the most popular dimensional reduction methods [2,3] to reduce the dimensions of the faces. This PCA technique can also be used in other pattern recognition problems such as speech recognition [4,5]. After using PCA to reduce the dimensions of the faces, the Eigenface use nearest-neighbor method [6] to recognize or to classify faces.

To classify the faces, a graph (i.e. kernel) which is the natural model of relationship between faces can also be employed. In this model, the nodes represent faces. The edges represent for the possible interactions between nodes. Then, machine learning methods such as Support Vector Machine [7], kernel ridge regression [8], Artificial Neural Networks [9] or the graph based semi-supervised learning methods [10,11,12] can be applied to this graph to classify or to recognize the faces. The Artificial Neural Networks, Support Vector Machine, and kernel ridge regression are supervised learning methods. The graph based semi-supervised learning methods are the semi-supervised learning methods. Please note that the kernel ridge regression method is the simplest form of the Support Vector Machine method.

In the last two decades, the SVM learning method has successfully been applied to some specific classification tasks such as digit recognition, text classification, and protein function prediction and face recognition problem [7]. However, the kernel ridge regression method (i.e. the simplest form of the SVM method) has not been applied to any practical applications. Hence in this paper, in addition to the nearest-neighbor method, we will also use the kernel ridge regression method applied to the face recognition problem.

Next, we will introduce the Principle Component Analysis. Principle Component Analysis (i.e. PCA) is one of the most popular dimensionality reduction techniques [2]. It has several applications in many areas such as pattern recognition, computer vision, statistics, and data analysis. It employs the eigenvectors of the covariance matrix of the feature data to project on a lower dimensional subspace. This will lead to the reduction of noises and redundant features in the data and the low time complexity of the nearest-neighbor and the kernel ridge regression approach solving face recognition problem.

However, the PCA has two major disadvantages which are the lack of sparsity of the loading vectors and each principle component is the linear combination of all variables. From data analysis viewpoint, sparsity is necessary for reduced computational time and better generalization performance. From modeling viewpoint, although the interpretability of linear combinations is usually easy for low dimensional data, it could become much harder when the number of variables becomes large. To overcome this hardness and to introduce sparsity, many methods have been proposed such as [13,14,15,16]. In this paper, we will introduce new approach for sparse PCA using Alternating Direction Method of Multipliers (i.e. ADMM method) [17]. Then, we will try to combine the sparse PCA dimensional reduction method with the nearest-neighbor method (and with the kernel ridge regression method) and apply these combinations to the face recognition problem. This work, to the best of our knowledge, has not been investigated.

Last but not least, we will also develop the novel tensor sparse PCA method and apply this dimensional reduction method to the face database (i.e. tensor) in order to reduce the dimensions of the face database. Then we will apply the nearest-neighbor method or the kernel ridge regression method to recognize the transformed faces.

We will organize the paper as follows: Section II will present the Alternating Direction Method of Multipliers. Section III will derive the sparse PCA method using the ADMM method in detail. Section IV will present the sparse PCA algorithm. Section V will present the detailed version of the tensor sparse PCA algorithm. In section VI, we will apply the combination of tensor sparse PCA algorithm with the nearest-neighbor algorithm and with the kernel ridge regression algorithm to faces in the dataset available from [18]. Section VII will conclude this paper and discuss the future directions of researches of this face recognition problem.

II. Alternating Direction Method of Multipliers

In this section, we will introduce the Alternating Direction Method of Multipliers. The detailed information about the Alternating Direction Method of Multipliers can be found in [17]. First, assume that we want to solve the following problem

$$minimize f(x) + g(z)$$

$$subject\ to\ Ax + Bz = c$$

with variables $x \in R^n$ and $z \in R^m$, where $A \in R^{p*n}, B \in R^{p*m}$.

Next, we will form the augmented Lagrangian

$$L_\rho(x,z,y) = f(x) + g(z) + y^T(Ax + Bz - c) + \frac{\rho}{2}||Ax + Bz - c||_2^2$$

Finally, $x^{k+1}, z^{k+1}$, and $y^{k+1}$ can be solved as the followings

$$x^{k+1} = argmin_x L_\rho(x, z^k, y^k)$$

$$z^{k+1} = argmin_z L_\rho(x^{k+1}, z, y^k)$$

$$y^{k+1} = y^k + \rho(Ax^{k+1} + Bz^{k+1} - c),$$

where $\rho > 0$.

III. Sparse Principle Component Analysis Derivation

Assume that we are given the data matrix $D \in R^{n*p}$ ($n$ is the number of samples and $p$ is the number of features). Next, we will formulate our sparse PCA problem. This problem is in fact the following optimization problem

$$minimize_{x,z} -||\tilde{D}x||_2^2 + \lambda||z||_1$$

$$such\ that\ x = z\ and\ ||z|| \leq 1.$$

Our objective is to find the sparse vector $x$. Please note that $\tilde{D} = \begin{bmatrix} d_1 - \mu \\ d_2 - \mu \\ \vdots \\ d_n - \mu \end{bmatrix}$, where $\mu = \frac{1}{n}\sum_{i=1}^n d_i$ be the mean vector of all row vectors $d_1, d_2, \ldots, d_n$ of $D$.

First, the augmented Lagrangian of the above optimization problem can be derived as the following

$$L_\rho(x,z,y) = -||\tilde{D}x||_2^2 + \lambda||z||_1 + y^T(x - z) + \frac{\rho}{2}||x - z||_2^2$$

Then $x^{k+1}, z^{k+1}, and\ y^{k+1}$ can be solved as the followings

$$x^{k+1} = argmin_x L_\rho(x, z^k, y^k)$$

Hence

$$\frac{dx^{k+1}}{dx} = \frac{d}{dx}(-||\tilde{D}x||_2^2 + y^{k^T}(x - z^k) + \frac{\rho}{2}||x - z^k||_2^2)$$

$$= \frac{d}{dx}(-x^T\tilde{D}^T\tilde{D}x + y^{k^T}(x - z^k) + \frac{\rho}{2}||x - z^k||_2^2)$$

$$= -2\tilde{D}^T\tilde{D}x + y^k + \rho(x - z^k))$$

Next, we solve $\frac{dx^{k+1}}{dx} = 0 \Leftrightarrow (-2\tilde{D}^T\tilde{D} + \rho I)x = -y^k + \rho z^k$

Thus, $x^{k+1} = (-2\tilde{D}^T\tilde{D} + \rho I)^{-1}(-y^k + \rho z^k)$

Next, we have

$$z^{k+1} = argmin_z L_\rho(x^{k+1}, z, y^k)$$

Hence

$$\frac{dz^{k+1}}{dz} = \frac{d}{dz}(\lambda||z||_1 + y^{k^T}(x^{k+1} - z) + \frac{\rho}{2}||x^{k+1} - z||_2^2)$$

$$= \lambda\xi - y^k + \rho(x^{k+1} - z)(-1)$$

$$= \lambda\xi - y^k + \rho(z - x^{k+1}),$$

where

$$\xi_i = \begin{cases} 1 \text{ if } z_i > 0 \\ [-1,1] \text{ if } z_i = 0 \\ -1 \text{ if } z_i < 0 \end{cases}$$

Solve $\frac{dz^{k+1}}{dz} = 0$, we have

$$z_i^{k+1} = x_i^{k+1} + \frac{1}{\rho}y_i^k - \frac{\lambda}{\rho}\xi_i$$

If $z_i^{k+1} > 0, \xi_i = 1$, then

$$x_i^{k+1} + \frac{1}{\rho}y_i^k - \frac{\lambda}{\rho} > 0 \Rightarrow x_i^{k+1} + \frac{1}{\rho}y_i^k > \frac{\lambda}{\rho}$$

If $z_i^{k+1} < 0, \xi_i = -1$, then

$$x_i^{k+1} + \frac{1}{\rho}y_i^k + \frac{\lambda}{\rho} < 0 \Rightarrow x_i^{k+1} + \frac{1}{\rho}y_i^k < -\frac{\lambda}{\rho}$$

If $z_i^{k+1} = 0$, then

$$-\frac{\lambda}{\rho} \leq x_i^{k+1} + \frac{1}{\rho}y_i^k \leq \frac{\lambda}{\rho}$$

Thus,

$$z_i = sign(x_i^{k+1} + \frac{1}{\rho}y_i^k)\max(|x_i^{k+1} + \frac{1}{\rho}y_i^k| - \frac{\lambda}{\rho}, 0)$$

Finally, we have

$$y^{k+1} = y^k + \rho(x^{k+1} - z^{k+1})$$

IV. Sparse Principle Component Analysis Algorithm
   In this section, we will present the sparse PCA algorithm

---

Algorithm 1: Sparse PCA algorithm

---

1. Input: The dataset $D \epsilon R^{n*p}$, where $p$ is the dimension of the dataset and $n$ is the total number of observations in the dataset

2. Compute $\widetilde{D} = \begin{bmatrix} d_1 - \mu \\ d_2 - \mu \\ \vdots \\ d_n - \mu \end{bmatrix}$, where $\mu = \frac{1}{n}\sum_{i=1}^{n} d_i$ be the mean vector of all row vectors $d_1, d_2, \ldots, d_n$ of $D$

3. Randomly select parameters $\rho, \lambda$.

4. Set $B = eye(p)$

5. Set $X = zeros(p, dim)$

6. for $i = 1: dim$

   i. Compute $L_\rho(x, z, y)$

   ii. Initialize $x^0, z^0, y^0$

   iii. Set $k = 0$

   iv. do

      a. Compute $x^{k+1} = argmin_x L_\rho(x, z^k, y^k)$

      b. Compute $z^{k+1} = argmin_z L_\rho(x^{k+1}, z, y^k)$

      c. Compute $y^{k+1} = y^k + \rho(x^{k+1} - z^{k+1})$

      d. $k = k + 1$

   v. while $||x^{k+1} - x^k|| > 10^{-10}$

   vi. $x = \frac{x^{k+1}}{norm(x^{k+1})}$

   vii. $X(:, i) = x$

   viii. $\widetilde{D} = \widetilde{D}(I - xx^T)$

7. End

8. Output: The matrix $\widetilde{D}X$.

---

V. Tensor Sparse Principle Component Analysis algorithm
   In this section, we will present tensor sparse PCA method.

Suppose $Y$ is a three-mode tensor with dimensionalities $N_1$, $N_2$, and $N_3$, where $N_1$ is the height of the image, $N_2$ is the width of the image, $N_3 = n_1 + \cdots + n_P$ is the total number of images in the dataset and $P$ is the total number of people in the dataset. Then, the first mode unfolding of $Y$ is written as $Y_{(1)}$ and is the matrix $N_2 N_3 * N_1$. The second mode unfolding of $Y$ is written as $Y_{(2)}$ and is the matrix $N_1 N_3 * N_2$. The third mode unfolding of $Y$ is written as $Y_{(3)}$ and is the matrix $N_1 N_2 * N_3$.

The refolding is the reverse operation of the unfolding operation. Thus the tensor sparse PCA can be represented as follows

---

Algorithm 2: Tensor Sparse PCA algorithm

---

1. Input: The three mode tensor $Y$
2. for $i = 1:2$
   i. Unfold the three mode tensor $Y$ to $Y_{(i)}$
   ii. Apply the sparse PCA algorithm (Algorithm 1) to $Y_{(i)}$ and get the output $Z_{(i)}$
   iii. Refold the matrix $Z_{(i)}$ to the three mode tensor $Y$
3. End
4. for $i = 1:P$
   i. Unfold the three mode sub-tensor $Y^{(i)}$ to $Y^i_{(3)}$
   ii. Apply the sparse PCA algorithm (Algorithm 1) to $Y^i_{(3)}$ and get the output $Z^i_{(3)}$
   iii. Refold the matrix $Z^i_{(3)}$ to the three mode sub-tensor $Y^{(i)}$
5. Merge all three mode sub-tensor $Y^{(i)}$ to form the final three mode tensor $Y$
6. Output: The tensor $C=Y$.

## VI. Experiments and Results

In this paper, the set of 120 face samples recorded of 15 different people (8 face samples per people) are used for training. Then another set of 45 face samples of these people are used for testing the accuracy measure. This dataset is available from [18]. Then, we will merge all rows of the face sample (i.e. the matrix) sequentially from the first row to the last row into a single big row which is the $R^{1*1024}$ row vector. These row vectors will be used as the feature vectors of the nearest-neighbor method and the kernel ridge regression method.

Next, the PCA and the sparse PCA algorithms will be applied to face samples in the training set and the testing set to reduce the dimensions of the face samples in the dataset. Then the nearest-neighbor method and the kernel ridge regression method will be applied to these new transformed feature vectors.

At the very beginning, each face sample is the $R^{32*32}$ matrix. There are 120 face samples in the training set. So, the training set is in fact the $R^{32*32*120}$ tensor. There are 45 face samples in the testing set. Hence the testing set is in fact the $R^{32*32*45}$ tensor. Then, we apply the tensor sparse PCA algorithm to the training set and the testing set to reduce the dimensions of the face dataset. Then, we will merge all rows of the face sample (i.e. the new matrix) into a single big row which is the $R^{1*625}$ row vector. Finally, the nearest-neighbor method and the kernel ridge regression method will be applied to the new transformed dataset.

In this section, we experiment with the above nearest-neighbor method and kernel ridge regression method in terms of accuracy measure. The accuracy measure Q is given as follows:

$$Q = \frac{True\ Positive + True\ Negative}{True\ Positive + True\ Negative + False\ Positive + False\ Negative}$$

All experiments were implemented in Matlab 6.5 on virtual machine. The accuracy performance measures of the above proposed methods are given in the following table 1 and table 2.

Table 1: **Accuracies** of the nearest-neighbor method, the combination of PCA method and the nearest-neighbor method, the combination of sparse PCA method and the nearest-neighbor method, and the combination of tensor sparse PCA method and the nearest-neighbor method

|  | Accuracy (%) |
|---|---|
| The nearest-neighbor method | 90.81 |
| PCA (d = 200) + The nearest-neighbor method | 91.11 |
| PCA (d = 300) + The nearest-neighbor method | 91.11 |
| PCA (d = 400) + The nearest-neighbor method | 91.11 |
| PCA (d = 500) + The nearest-neighbor method | 91.11 |
| PCA (d = 600) + The nearest-neighbor method | 91.11 |
| Sparse PCA (d = 200) + The nearest-neighbor method | 91.70 |
| Sparse PCA (d = 300) + The nearest-neighbor method | 91.70 |
| Sparse PCA (d = 400) + The nearest-neighbor method | 91.41 |
| Sparse PCA (d = 500) + The nearest-neighbor method | 91.70 |
| Sparse PCA (d = 600) + The nearest-neighbor method | 91.11 |
| **Tensor sparse PCA + The nearest-neighbor method** | **92** |

Table 2: **Accuracies** of the kernel ridge regression method, the combination of PCA method and the kernel ridge regression method, the combination of sparse PCA method and the kernel ridge regression method, and the combination of tensor sparse PCA method and the kernel ridge regression method

|  | Accuracy (%) |
|---|---|
| The kernel ridge regression method | 95.85 |
| PCA (d = 200) + The kernel ridge regression method | 96.15 |
| PCA (d = 300) + The kernel ridge regression method | 96.15 |
| PCA (d = 400) + The kernel ridge regression method | 96.15 |
| PCA (d = 500) + The kernel ridge regression method | 96.15 |
| PCA (d = 600) + The kernel ridge regression method | 96.15 |
| Sparse PCA (d = 200) + The kernel ridge regression method | 96.15 |
| Sparse PCA (d = 300) + The kernel ridge regression method | 96.44 |
| Sparse PCA (d = 400) + The kernel ridge regression method | 96.44 |
| Sparse PCA (d = 500) + The kernel ridge regression method | 96.15 |
| **Sparse PCA (d = 600) + The kernel ridge regression method** | **96.74** |
| Tensor sparse PCA + The kernel ridge regression method | 87.85 |

From the above tables, we recognize that the combination of the tensor sparse PCA method and one specific classification system does not always reach the best accuracy performance measure. This accuracy depends on the classification system. However, the accuracy of the combination of the sparse PCA method and one specific classification system is always better than the accuracy of the combination of the PCA method and one specific classification system and is always better than the accuracy of the classification system itself.

VII.     Conclusions

In this paper, the detailed versions of the sparse PCA method and the tensor sparse PCA method has been proposed. The experimental results show that the combination of the tensor sparse PCA with any specific classification system does not always reach the best accuracy performance measure. However, the accuracy of the combination of the sparse PCA method and one specific classification system is always better than the accuracy of the combination of the PCA method and one specific classification system and is always better than the accuracy of the classification system itself.

In the future, we will test the accuracies of the combination of the tensor sparse PCA method with a lot of classification systems, for e.g. the SVM method or deep neural network methods.

In this paper, we use the Alternating Direction Method of Multipliers (i.e. the ADMM method) to solve the sparse PCA problem. In the near future, we will use the other optimization methods such as the proximal gradient method to solve the sparse PCA problem. Then, we will compare the accuracies and the run time complexities among these optimization techniques. To the best of our knowledge, this work has not been investigated up to now.

Acknowledgment: This research is funded by Ho Chi Minh City University of Technology – VNU-HCM under grant number T-ĐĐT-2018-79.

References

1. Turk, Matthew, and Alex Pentland. "Eigenfaces for recognition." *Journal of cognitive neuroscience* 3.1 (1991): 71-86.
2. Tran, Loc Hoang, Linh Hoang Tran, and Hoang Trang. "Combinatorial and Random Walk Hypergraph Laplacian Eigenmaps." *International Journal of Machine Learning and Computing* 5.6 (2015): 462.
3. Tran, Loc, et al. "WEIGHTED UN-NORMALIZED HYPERGRAPH LAPLACIAN EIGENMAPS FOR CLASSIFICATION PROBLEMS." *International Journal of Advances in Soft Computing & Its Applications* 10.3 (2018).
4. Trang, Hoang, Tran Hoang Loc, and Huynh Bui Hoang Nam. "Proposed combination of PCA and MFCC feature extraction in speech recognition system." *2014 International Conference on Advanced Technologies for Communications (ATC 2014)*. IEEE, 2014.
5. Tran, Loc Hoang, and Linh Hoang Tran. "The combination of Sparse Principle Component Analysis and Kernel Ridge Regression methods applied to speech recognition problem." *International Journal of Advances in Soft Computing & Its Applications* 10.2 (2018).
6. Zhang, Zhongheng. "Introduction to machine learning: k-nearest neighbors." *Annals of translational medicine* 4.11 (2016).
7. Scholkopf, Bernhard, and Alexander J. Smola. *Learning with kernels: support vector machines, regularization, optimization, and beyond*. MIT press, 2001.
8. Trang, Hoang, and Loc Tran. "Kernel ridge regression method applied to speech recognition problem: A novel approach." *2014 International Conference on Advanced Technologies for Communications (ATC 2014)*. IEEE, 2014.
9. Zurada, Jacek M. *Introduction to artificial neural systems*. Vol. 8. St. Paul: West publishing company, 1992.
10. Tran, Loc. "Application of three graph Laplacian based semi-supervised learning methods to protein function prediction problem." *arXiv preprint arXiv:1211.4289* (2012).
11. Tran, Loc, and Linh Tran. "The Un-normalized Graph p-Laplacian based Semi-supervised Learning Method and Speech Recognition Problem." *International Journal of Advances in Soft Computing & Its Applications* 9.1 (2017).
12. Trang, Hoang, and Loc Hoang Tran. "Graph Based Semi-supervised Learning Methods Applied to Speech Recognition Problem." *International Conference on Nature of Computation and Communication*. Springer, Cham, 2014.

13. R.E. Hausman. Constrained multivariate analysis. Studies in the Management Sciences, 19 (1982), pp. 137–151
14. Vines, S. K. "Simple principal components." *Journal of the Royal Statistical Society: Series C (Applied Statistics)* 49.4 (2000): 441-451.
15. Jolliffe, Ian T., Nickolay T. Trendafilov, and Mudassir Uddin. "A modified principal component technique based on the LASSO." *Journal of computational and Graphical Statistics* 12.3 (2003): 531-547.
16. Zou, Hui, Trevor Hastie, and Robert Tibshirani. "Sparse principal component analysis." *Journal of computational and graphical statistics* 15.2 (2006): 265-286.
17. Boyd, Stephen, et al. "Distributed optimization and statistical learning via the alternating direction method of multipliers." *Foundations and Trends® in Machine learning* 3.1 (2011): 1-122.
18. http://www.cad.zju.edu.cn/home/dengcai/Data/FaceData.html